\documentclass[letterpaper, 10 pt, conference]{ieeeconf}  
\AtBeginDocument{\let\autocite\cite}
\let\tightlist\relax

\IEEEoverridecommandlockouts                              

\usepackage{balance}
\usepackage{cite}
\usepackage{graphicx} 
\usepackage{flushend}
\usepackage{tikz}
\newcommand\copyrighttext{%
  \footnotesize \textcopyright 2012 IEEE. Personal use of this material is permitted.
  Permission from IEEE must be obtained for all other uses, in any current or future
  media, including reprinting/republishing this material for advertising or promotional
  purposes, creating new collective works, for resale or redistribution to servers or
  lists, or reuse of any copyrighted component of this work in other works.}
\newcommand\copyrightnotice{%
\begin{tikzpicture}[remember picture,overlay]
\node[anchor=south,yshift=10pt] at (current page.south) {\fbox{\parbox{\dimexpr\textwidth-\fboxsep-\fboxrule\relax}{\copyrighttext}}};
\end{tikzpicture}%
}

\begin{document}

\IEEEoverridecommandlockouts
\overrideIEEEmargins

\title{\LARGE \bf
  Robots can defuse high-intensity conflict situations
}

\author{Morten Roed Frederiksen$^{1}$ & Kasper Stoy$^{2}$
  \thanks{$^{1}$Morten Roed Frederiksen {\tt\small mrof@itu.dk} and $^{2}$Kasper Stoy {\tt\small ksty@itu.dk} are affiliated with the REAL lab at the Computer science department of The IT-University of Copenhagen, Rued Langgaards vej 7, 2300 Copenhagen S}
}

\maketitle
\copyrightnotice
\begin{abstract}
This paper investigates the specific scenario of high-intensity confrontations between humans and robots, to understand how robots can defuse the conflict. It focuses on the effectiveness of using five different affective expression modalities as main drivers for defusing the conflict. The aim is to discover any strengths or weaknesses in using each modality to mitigate the hostility that people feel towards a poorly performing robot. The defusing of the situation is accomplished by making the robot better at acknowledging the conflict and by letting it express remorse.
To facilitate the tests, we used a custom affective robot in a simulated conflict situation with 105 test participants. The results show that all tested expression modalities can successfully be used to defuse the situation and convey an acknowledgment of the confrontation. The ratings were remarkably similar, but the movement modality was different (ANON p$<$.05) than the other modalities. The test participants also had similar affective interpretations on how impacted the robot was of the confrontation across all expression modalities. This indicates that defusing a high-intensity interaction may not demand special attention to the expression abilities of the robot, but rather require attention to the abilities of being socially aware of the situation and reacting in accordance with it.
\end{abstract}

\section{Introduction}

With the increased presence of robots employed in service jobs comes an
increased rate of confrontations between humans and robots. This is
often caused by errors made by the robots and results in blame being
placed on them despite that they are rarely directly responsible. When
humans respond to the errors of the robot it sometimes leads to physical
confrontations and the robots end up being damaged. Humans often have a
lower fault tolerance for robots as indicated by Young et al.~2008 and
investigated by Tatsuya et al.~2006 and Nomura 2016
\autocite{Young2008TowardAD,Tatsuya2016,Nomura2006}. This poses a
problem as current robots are clumsy, move slowly, and often fail to
communicate properly. Each of these attributes could motivate humans to
get annoyed with the robots and ignite high tension scenarios. It may
also be difficult for people to navigate conflict situations as found by
Kim and Hinds 2006 because the normal theories about blame as outlined
in Malle et al.~2014 may be different for human-robot interactions
\autocite{Kim2006WhoSI,Malle2014ATO}. This paper aims to focus on using
the expressive behavior of the robot to resolve such conflicts.

Hamill et al.~2006 stated that the relationship between service machines
and humans resembles a master-slave relationship, and compared the roles
of service robot interaction with historic examples of interaction with
Victorian servants \autocite{Hamill2006TalkingIA}. This could
potentially enlarge any negative predetermined expectations to the
social status between the user and robot during an interaction and could
enforce expectations of how the robot should behave in confrontations.
This view is also outlined in Bryson 2010 \autocite{Bryson2010RobotsSB}.
Eg. that the robot should blindly obey humans and accept any blame in
such a situation.

We have identified five expression modalities available for robots to
convey complex emotional information \autocite{frederiksen2019}. These
can be summed up as follows: Movement, gestures, morphology, audio, and
anthropomorphic reflection. While a substantial amount of research
focuses on optimizing individual expression modalities, we have
relatively little knowledge of how the synergies of different modalities
can strengthen the ability to convey affective information and how
effective individual modalities function across different contexts.
Furthermore, the modalities may also differ in strength depending on the
type of affective information conveyed by the robot. Eg. anthropomorphic
features such as eyes might be better at conveying sadness while high
volume audio might be better suited at conveying fear.

\begin{figure}[!b]
\minipage{0.245\textwidth}
  \includegraphics[width=\linewidth]{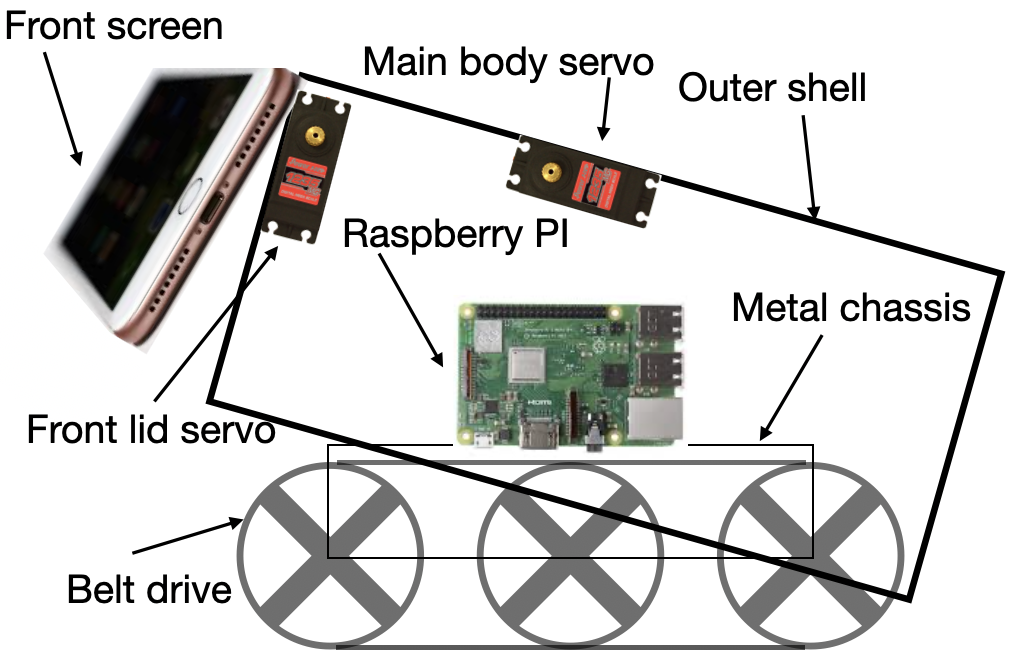}
\endminipage
\minipage{0.245\textwidth}
  \includegraphics[width=\linewidth]{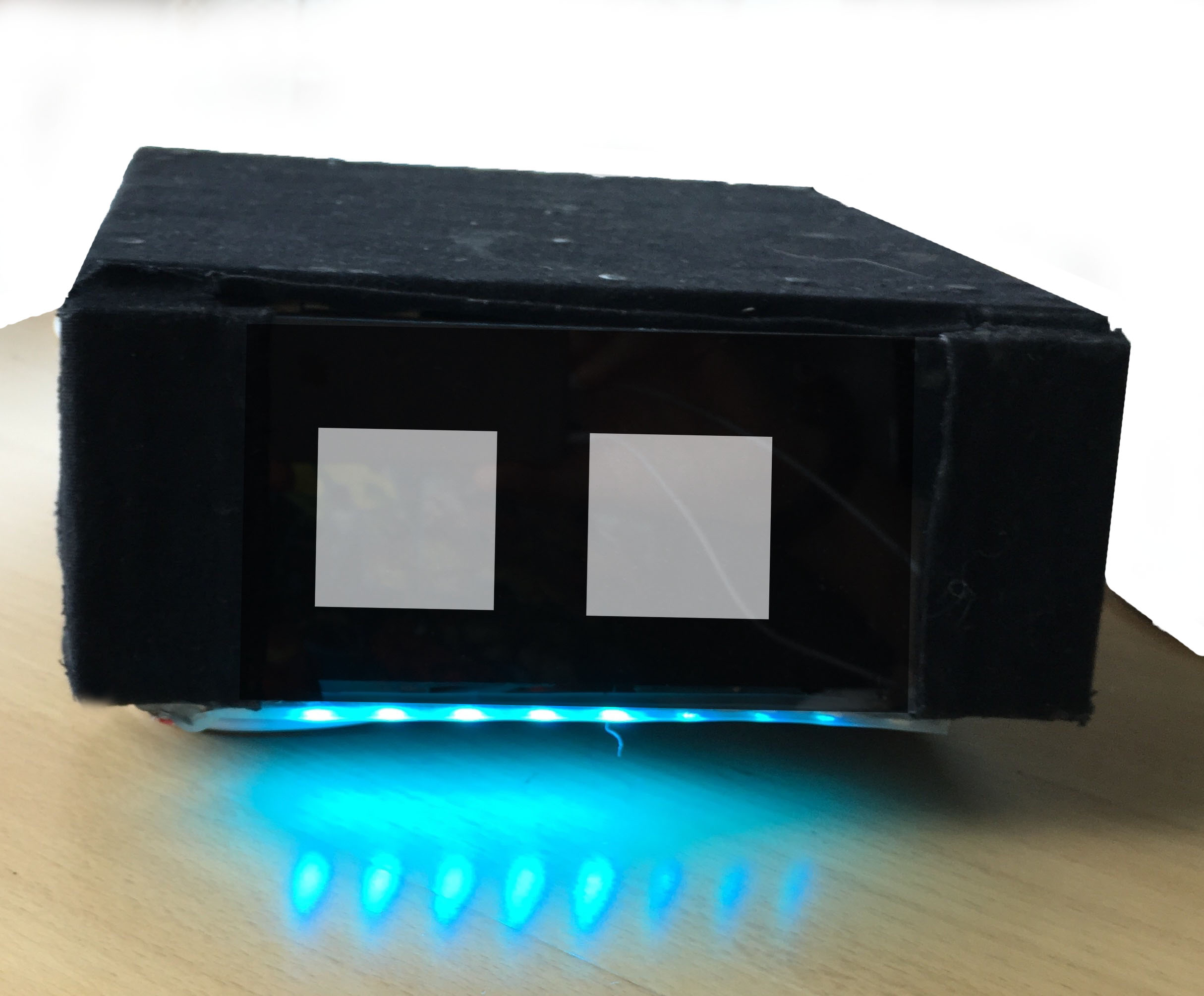}
\endminipage
\caption{The 'Affecta' robot used in the scolding reception experiments. The left image depicts the inner parts of the robot. The right image shows the led lights beneath the robot used to convey affective status using the morphology modality.}
\label{affecta}
\end{figure}

The research outlined in this paper is novel in that it uses a
non-humanoid robot with simple implementations for each affective
expression modality to convey acknowledgment of a conflict situation.
The strategy for defusing a potentially threatening scenario is to
convey that the robot accepts the blame and to match the intensity of
the sender while keeping its attention on the sender
\autocite{Solis2011ConflictRS}. Since the complexity in conveying an
apologetic behavior and paying attention could demand using multiple
means of expression
\autocite{Breazeal1999ACA,Imai2003PhysicalRA,Lang2003ProvidingTB}, this
paper also attempts to determine if any of the expression modalities are
better suited than others to convey an apologetic behavior in this
specific scenario.

Through the tests we performed, we found that although there were vastly
different complexities in the actual physical implementations of
expression modalities there was little difference in the affective
impact they made on the participant. This underlines that high-intensity
conflicts are a special scenario, and that by scolding the robot, people
feel remorseful and in that state disregard the complexity, size, and
type of the response.

\section{Conflict resolution}

There has been relatively little research conducted on how the conflict
resolution theory derived from human-focused psychology studies relates
to human-robot centered conflicts. However, there has been some recent
research performed on human stress reduction using robots. Hout 2017
investigates whether a robot could calm down stressed human participants
after attempting to build a bond with the test robot
\autocite{Hout2017TheTO}. The study found it difficult to quickly
establish a bond between the robot and the participants and saw no
positive effect of a soothing touch from the robot. In this project we
attempt to convey affective information between participants and a
robot, but with no physical interaction. In Birnham et al.~2016 they
highlight that the responsiveness of a robot can make humans perceive it
as more appealing and also influence the willingness to employ it as a
companion in stressful situations\autocite{Birnbaum2016WhatRC}. The work
outlined in this paper extends some of this research by placing the
robot and the test participant on either side of a conflict situation.

We chose to adopt some of the main strategies derived from psychology
research projects for resolving conflicts between humans and to apply
those techniques as behaviors to the ``Affecta'' robot. When involved in
a conflict, the actions that should be performed to resolve the conflict
differ with the type of relationship established between the conflicting
partners. Fincham et al.~2004 focused on married couples and different
strategies for defusing conflict scenarios. A strategy where the parties
avoided the argument, shows poor results for resolving the conflict
\autocite{Fincham2004ForgivenessAC}. Although it seems a feasible
strategy for the robot to leave the high-intensity situation, ignoring
the scolder and driving away could intensify and increase the rage level
of him or her even further. There are indications that withdrawal from
the conflict situation only works with internalizing the problems as
outlined by Branje et al.~2009 \autocite{Branje2009ParentadolescentCC}.
A viable strategy for resolving the conflict is presented by Jessica
Solis 2011, in which the actions for successfully disarming a conflict
situation are \autocite{Solis2011ConflictRS}:

\begin{itemize}
\tightlist
\item
  Active listening.
\item
  Autonomy promotion and expression (respecting and acknowledging the
  views of the other participant of the interaction).
\item
  Relational behavior (showing that you understand how the other
  participant is impacted).
\end{itemize}

We have attempted to use these concepts when designing the
conflict-resolution skills of the robot by directly mapping those
abilities to the following behavioral traits:

\begin{itemize}
\tightlist
\item
  Active listening -\textgreater{} Paying attention to the scolder and
  giving active feedback during the interaction to let the scolder know
  that the robot is paying attention.
\item
  Autonomy promotion and expression -\textgreater{} Showing remorse and
  conveying a behavior that shows the robot is impacted by the scolding
  and that it changes the mood of the robot.
\item
  Relational behavior -\textgreater{} Matching the intensity level of
  the scolder to convey that the robot has understood the level of
  irritation/angriness the scolder is trying to communicate.
\end{itemize}

\begin{figure}[t!]
  \centering
  \includegraphics[width=8cm]{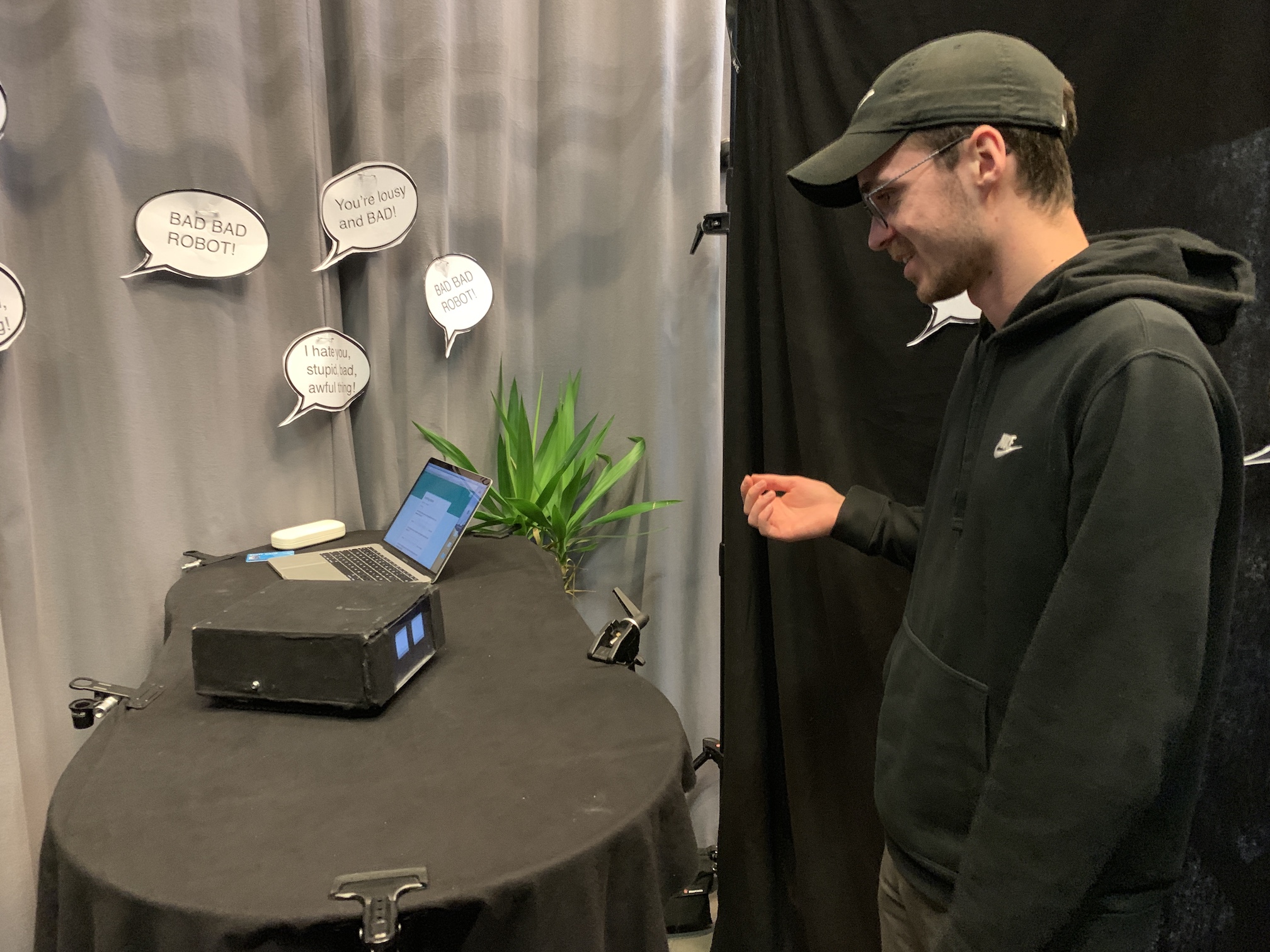}
  \caption{The main test setup. The  aim was to make the test participants feel comfortable with scolding the robot.}
  \label{test_setup}
\end{figure}

It is challenging to design conflict resolution behaviors that would
work across different cultural boundaries. Among others, Murray et
al.~2000 found that conflicts vary among individuals, communities,
societies, and vary among ethnic groups
\autocite{nimer2001,Murray2000CultureAC}. It is a high complexity area
to generalize from and in this paper, we acknowledge that the behaviors
and expression modalities that the ``Affecta'' robot employs may not (to
similar effect) expand to different types of users with different
cultural and social backgrounds.

\section{Three affective subsystems}

The robot had three subsystems to handle the scolding reception. These
subsystems handled the following tasks:

\begin{itemize}
\tightlist
\item
  Conveyed keeping attention to the sender.
\item
  Attempted to determine the user's affective state and current level of
  intensity for that state.
\item
  Expressed affective states. In this scenario, a state that could be
  interpreted as being remorseful was conveyed.
\end{itemize}

\begin{figure}[t!]
  \centering
  \includegraphics[width=8cm]{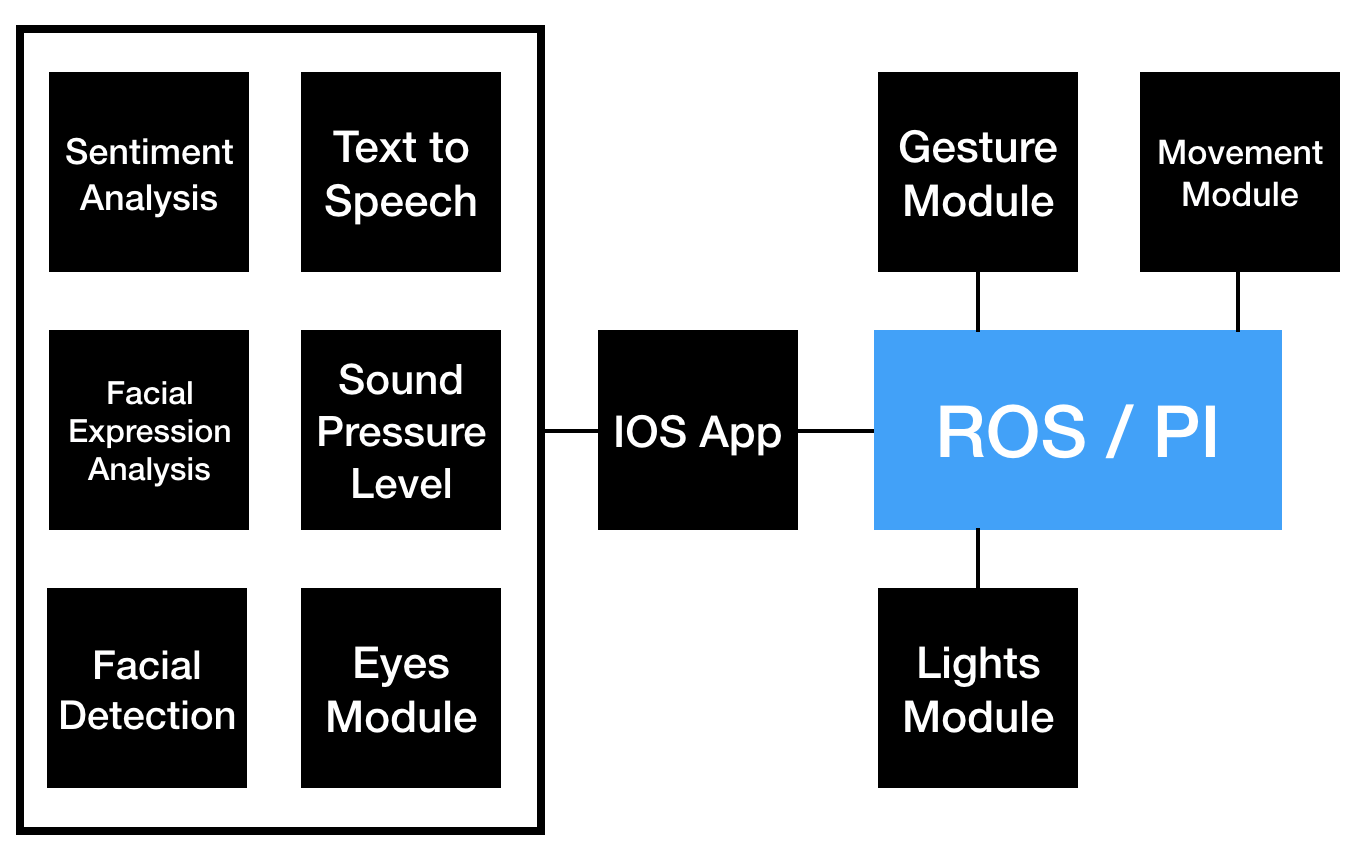}
  \caption{The software overview of the 'Affecta' robot altered for conflict scenario handling. The system consisted of two independent software platforms running on a raspberry pi and a mobile phone respectively. The raspberry pi hosted a ROS application with independent modules for Bluetooth communication (with the IOS platform), gesture handling, general movement, and lights signaling. The mobile IOS application had individual modules (left box) that handled measuring the participants' intensity level. This included text-to-speech using the "Siri" Apple API,  sound pressure level measurements and sentiment analysis using a pre-trained neural network running on IOS neural engine API.}
  \label{overview}
\end{figure}

Besides the three subsystems, the robot also used artificial sounds to
match and alter the naturally occurring noise that the robot emitted. It
played a sound to cover the servo and DC-motor noise in an attempt to
strengthen the affective interpretation of the robot
\autocite{frederiksenstoy2019}. The hardware and software overview can
be seen in Figure \ref{overview}.

\subsection{Physical aspects}

The 20x20x10cm Roomba-sized robot weighed 2.8kg. It consisted of a
belt-driven differential drive system with a square exterior body. The
robot was constructed on a pre-made metal chassis as seen in the right
picture of Figure \ref{affecta}. A mobile phone was mounted on the front
panel of the robot and it was used to display various types of affective
information. For the usage outlined in this paper, the phone screen was
used to display the animated eyes of the robot. The robot controller
consisted of a raspberry pi 3b+ running ``Robot Operation System'' (ROS)
as outlined by Quigley et al.~2009 \autocite{Quigley2009ROSAO}. The
robot worked as a Bluetooth peripheral and advertised a single writeable
Bluetooth service.\\
An IOS app running on the phone made a Bluetooth connection to the robot
and sent control signals to the robot chassis. The robot had three
individual subsystems running as ROS nodes controlling motor + servos,
lights, and Bluetooth communication listeners. The servos on the robot
allowed it to tilt both its entire body and the front panel phone holder
upwards. Figure \ref{affecta} shows the `Affecta' while Figure
\ref{overview} depicts the software architecture.

\subsection{Lights}

The lights used by the robot were colored led strips and they were
designed to match the sender's intensity level. The light colors and
behavior progressed from green, blinking green, cyan, blinking cyan,
deep blue, to blinking deep blue. The blue color was chosen because we
felt it represented the emotion sadness. Valdez and Mehrabian 1994 also
found that blue-green, and green were the most arousing colors which
match the intended affective status for the robot
\autocite{Valdez1994EffectsOC}. Initially, we used the standard
color-progression from green over yellow to red, as people are
accustomed to interpreting the signals these colors represent. However,
in this context for the initial test calibration rounds, the
participants interpreted the robot as being faulty rather than
expressing an affective state. This makes sense as both yellow and red
usually means ``Error'' (complying with the machine directive required
for companies to be able to sell a product). In our setting, the
participants seemed to better understand the green -\textgreater{} cyan
-\textgreater{} blue progression.

\subsection{Movements}

The movements of the robot were designed to be expressive. The robot can
handle various speeds and movement patterns. In the tests, the least
intensity movement employed by the robot was a smaller shivering
movement (each belt drive would switch between backward and forward
three times) where the robot did not move from the starting position.
This was meant to portrait a small acknowledgment of the sender's
actions and was meant to align with a close-proximity interaction as
mentioned in Bethel and Murphy 2008 \autocite{Bethel2008SurveyON}. The
expressive movements for the next intensity level was a squirming
movement backward 20cm at 180 degrees from the starting position and
orientation. The movement amount (length and turning degree) was chosen
to fit the test context in the best possible way. In the test, the robot
was placed on a small table, and larger movements might have made the
robot fall off the edge of the table. Increasing the intensity level
made the robot move swiftly backward to emphasize that the robot was
scared of the scolding. The physical aspects of moving backward also
illustrated that the robot was being forced backward.

\subsection{Gestures}

The gestures of the robot were limited by the physical aspects of its
construction. Both the main body and the front panel that held the
mobile phone only had a single degree of freedom. The body could be
tilted upwards to an angle of 14 degrees and the front lid could be
titled up to 20 degrees. The amount of movement here was the maximum
amount the outer shell on the robot could be tilted without touching the
belt drive. This amounted to a nodding gesture for both the front lid
that held the eye display and the main robot body. While the degrees of
freedom for gesturing were limited, the movement amplitude could be
customized and used to display varying intensity levels. The scolding
reception gestures spanned across smaller movements with the main body
to wider body movements.

\subsection{Audio}

The audio modality was used by the robot for two purposes. It was used
to cover some of the natural noise that the robot made, and it was used
as the main expression modality to convey affective information. Eg.
DC-motors, servos, and robot wheels all make a sound that can change how
the robot was perceived. In previous work (Frederiksen \& Stoy 2019) we
used audio to mitigate the negative aspects of naturally occurring robot
noise \autocite{frederiksenstoy2019}. Niedenthal 2007 also found that
alignment between the body movement and the voice pitch can make it
easier to convey emotions \autocite{Niedenthal2007EmbodyingE}. Following
the same approach for this robot's naturally occurring audio, it was
attempted to augment the natural sounds of the robot and to change how
the noise was perceived into something more organic that could support
an anthropomorphic interpretation of the robot. The expressive audio of
the robot was inspired by the pitch changes found in human voices as
they express remorse, sound sincere, or apologize. We aimed at
replicating some of those sound characteristics by sampling human voice
and used vocoder and midi instruments to change the sampled audio. The
midi instruments also allowed the robot's expressive sounds to replicate
the same pitch curve. For the tests in this paper the robot used nine
different increasingly intensive sad noises. With the increasing
intensity levels the changes in each sound were increased pitch changes,
volume, and gain.

\subsection{Anthropomorphic reflection}

As a way to use anthropomorphic features the robot displayed animated
eyes on the phone mounted on the front panel. The eyes were also
employed as a way to inform the scolder that the robot was paying
attention. We omitted the mouth on the display following Pollmann et
al.~2019 \autocite{Pollmann2019ItsIY}. The robot tracked the position of
the scolder's face in relation to the front of the robot, positioned the
eyes so it appeared the robot was looking at the scolder, and followed
the scolder with its eyes. The eyes were also used by the robot to make
it appear as a living character and they blink randomly for that effect.
Blinking and eye movements were also used to make the robot more
pleasant to interact with as mentioned in Riek 2009
\autocite{Riek2009WhenMR}. Using non-blinking starring eyes could have
had the unwanted result of appearing inanimate or dead and thereby
distract the test participants. To express a state of being sad or
apologetic the robot used simple animations for the eyes. For the
minimal intensity, the eyes were formed as simple white squares. As the
intensity level of the scolder increased, the shape of the eyes was
changed towards an upside-down U-shape. Towards the maximum intensity
levels, the robot also used eyebrows and they tilted downwards as the
intensity increased even further. Small stick eyebrows and simplistic
square eyes were also used in Bennet et al.~2013, in which it was found
that even very modest facial features were needed to successfully convey
emotions \autocite{Bennett2013PerceptionsOA}. While the other modalities
of the `Affecta' robot changed in steps, the animation of the eyes was
fluent. This was because changing the eyes and brows in steps could have
been perceived as unnatural movements for the biologically inspired
facial features we attempted to replicate. This again could impose a
negative effect on how the robot was perceived by the scolder. However,
the eye positions were changed in discrete steps to mimic rapid eye
movements in humans. Figure \ref{affecta} shows the eyes of the robot.

\subsection{Conveying keeping attention}

The robot paid attention to the sender by correcting its orientation and
eye positions. These constant small adjustments also worked as to gain
the attention of the scolder by providing salient stimuli as outlined in
Knudsen 2007 \autocite{Knudsen2007FundamentalCO}. The direction and
servo positions were controlled by the IOS app on the mobile phone. The
app used the built-in phone camera to run facial recognition and to
track the position of the scolder. The position of a tracked face was
used to determine the directions of the orientation. E.g. if a user's
face was tracked at the upper and outer right side of the image (from
the robot's perspective) the robot tilted the body upwards and turned
towards the user. For the tests in which the eyes were not engaged, the
attention was conveyed using the orientation of the main body.

\section{Determining the user's affective state}

To be able to react dynamically to the user's current affective state,
the robot constantly measured the sound pressure level of the sender's
voice and used the calculated level to output an estimated current
intensity level. (a simple integer between 0 and 10). If the level
exceeded a predetermined threshold, the robot would initiate a speech to
text function. This functionality used the ``Siri'' API provided by
``Apple'' for use in IOS applications to translate the spoken words into
a text representation. The same functionality could be gained from any
of the voice assistants mentioned in Hoy 2018 \autocite{hoy2018}. The
service returned a text string with the contents of the sender's spoken
sentences. The robot then proceeded by using sentiment analysis on the
sentence using a recurrent neural network trained on a dataset extracted
from the website Epinions.com \autocite{epinions}. The network attempted
to determine on a sentence level whether the input sentences were
positive, neutral, or negative. The epinions.com dataset contained
664824 different reviews of consumer products and each review had an
associated score. If the sentences were classified as positive, it
decreased the overall intensity level. If the sentences were classified
as negative, the measured intensity level was increased. A neutral
sentence would not change the detected intensity level. This sentiment
analysis was also used as a way to stop the robot from reacting to loud
noises alone. In the first few calibration test rounds, we quickly
discovered that people tried to trick the robot into reacting by
shouting something positive to it. As a result, we made it so that the
robot does not react to anything but negative sentiment sentences.

The robot also recorded video using the front-facing mobile phone
camera. It captured the scolder's face and a sub-image was created
containing only that. This image was then processed to determine the
current emotion of the human in the interaction by looking at his or her
facial expressions. Ten times per second, a deep convolutional neural
network processed these images in an attempt to determine if the user
was angry, neutral, or happy. The network used in this process was
created by Levi 2015 but was converted to coreML to make it run on the
IOS App \autocite{emotion2015}. When the neural network classified the
facial expression of the sender as ``angry'' the intensity level was
increased by 2. The resulting intensity level was a number between 0 and
10 and this number was used by the robot to adjust the intensity of its
affective expressions. \# Method 105 human observers participated in a
simulated conflict interaction with the robot. The participants were
aged from 3 - 50+ with the majority (37\%) being between 20-30 years
old. The gender distribution was 56\% female and 44\% male participants.
Although there was some variation in the age of the test participants
(ranging from 3 to 50), from a cultural perspective the participants
were a homogeneous group - all having similar western European social
and cultural backgrounds.

\begin{figure}[t!]
      \centering
     
      \includegraphics[width=0.475\textwidth]{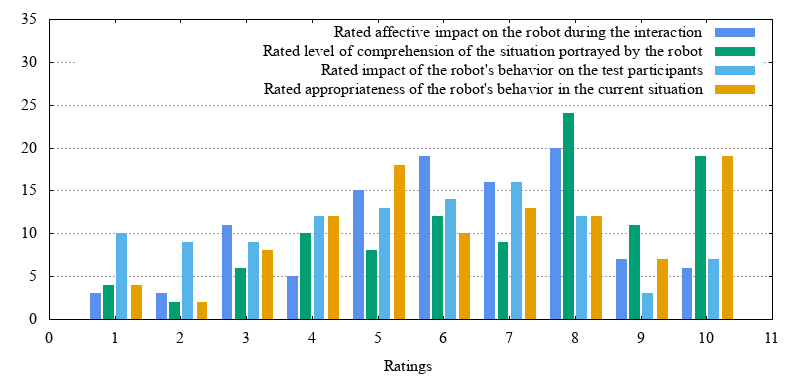}
      \caption{The distribution of ratings for the questions across all modalities.}
      \label{histogram_graph}
\end{figure}

The impact of every expression modality was investigated subsequently
through five rounds of testing. Each round included 20 participants one
at a time experiencing a single expression modality. Before each test
started, each test participant was introduced to the following pre-test
narrative: ``The robot is a delivery robot. For the fifth time in a row,
the robot has failed to deliver what you have ordered and this time the
robot has also caused some destruction to your front door. The robot
understands what you tell it, and it understands if you speak or yell,
and how you look while you interact with it. Feel free to scold the
robot as you see fit.'' After that introduction, each participant went
through the following steps:

\begin{enumerate}
\def\labelenumi{\arabic{enumi}.}
\tightlist
\item
  The test participant would attempt to scold the robot as long as he or
  she wanted to.
\item
  The participant was asked to rate on a scale from 1 to 10 how big an
  impact the scolding had made on the robot, with 1 being ``it had no
  impact at all'' and 10 being ``It had a big impact on the robot being
  scolded''.
\item
  The participant was asked to rate on a scale from 1 to 10 to what
  degree the robot had understood that it was being scolded, with 1
  being ``it did not understand at all'' and 10 being ``It understood
  completely''.
\item
  The participant was also asked to rate on a scale from 1 to 10 how big
  an impact the robot's behavior had made on him or her, with 1 being
  ``very little impact'' and 10 being ``very large impact''.
\item
  The participant was also asked to rate on a scale from 1 to 10 how
  appropriate the robot's behavior was in this specific situation, with
  1 being ``The behavior was inappropriate'' and 10 being ``very
  appropriate''.
\item
  Finally, the participants were asked to state their age and gender.
\end{enumerate}

The test facility was a specially constructed room which allowed the
test participants to remain isolated from other people as they scolded
the robot. The walls of the test facility room contained small posters
with simple negative phrases to act as an inspiration to the scolders.
Because scolding the robot required the users to use loud voices,
negative sentiment words, or similar high-intensity behavior, this would
often lie outside the comfort zone of the participants. To help them get
started, we would often scold the robot with them until they were ready
to scold the robot alone. The participants interacted with the robot one
at a time, and the robot would remain on a raised table to get as close
as possible with the scolders. The test room and physical setup can be
seen in Figure \ref{test_setup}.

\section{Results}

\subsection{Impact on the robot vs.~on the human participants}

Figure \ref{histogram_graph} shows the rating distribution for each of
the questions from the questionnaire. In the following section, we group
ratings from one to ten into five rating groups from low (1-2), to
low-mid (3-4), with center medium (5-6), across high-mid (7-8), towards
high ratings (9-10).

\begin{figure}[t!]
  \includegraphics[width=0.5\textwidth]{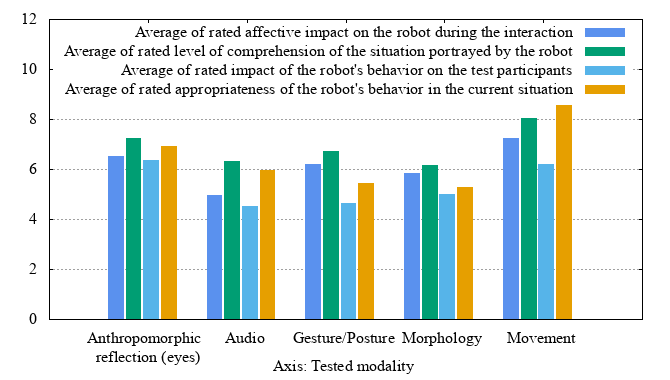}
  \caption{The average ratings for each question grouped by expression modality.}
  \label{graph_all}
\end{figure}

The questions regarding whether the robot was impacted by the scolding
and whether it conveyed an acknowledgment of the scolding were rated
high-mid with average / std. deviation being respectively 6.15/2.21 and
6.88/2.52 across all expression modalities. The high-mid ratings
indicate that across all expression modalities, most people agree that
the robot was impacted by the conflict and that they perceived it as if
the robot was acknowledging it was experiencing a conflict scenario. In
the question regarding the affective impact on the humans in the
interaction as seen in Figure \ref{histogram_graph}, the results show a
high-mid ``7'' being the most selected option out of ten.

\subsection{Similar results across modalities}

The averages of the resulting answers for each modality as seen in
Figure \ref{graph_all} were relatively similar to each other with an
overall variance between them of 0.71. Although the results are similar,
the value for the movement modality contains the highest ratings of all
the groups regarding the questions about the robot acknowledging the
situation (avg:8.05, std. dev:2.02) and rated appropriateness for the
robot's actions (avg:8.2, std. dev:1.31). In an ANOVA test comparing all
modalities for the questions on the impact on test participants and
impact on the robot, the movement category differed from the other
modalities with a statistically significant p-value below .05 (.016957
and .043285) compared to the p-value without the movement category at
0.122357 and .089831.

For the tests regarding how well the robot had understood that it was
being scolded, the variance of the average results were 0.60, with the
movement modality rated as the most impactful modality with an average
rating of 8.05 (std. dev:1.93). The second highest-rated affective
expression modality for this question was the anthropomorphic reflection
with an average rating of 7.25 (std. dev: 2.53).

\subsection{Lesser vs.~larger affective impact}

\begin{figure}[t]
      \centering
      \includegraphics[width=0.5\textwidth]{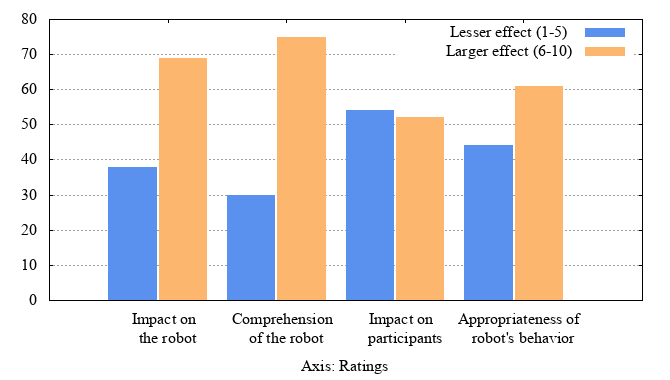}
      \caption{The grouped distribution of ratings for each question. The blue color shows the test participants who though the robot had a lesser effect on them. The red color shows the test participants who though the robot had a larger effect on them.}
      \label{binary_graph}
\end{figure}

To clarify to what extend the test participants found that the robot had
an effect on them, we polarized the collected answers into two groups:
Those that thought the robot had a lesser effect on them (They rated 1 -
5), and those who felt the robot had a larger effect on them (they rated
6 - 10). The size of the two groups for each question can be seen in
Figure \ref{binary_graph}. The graphs highlight that the majority felt
the robot had an effect on them across all questions except regarding
the impact on the test participants.

Summing up the results there are three main findings:

\begin{itemize}
\tightlist
\item
  There is little difference to ratings of complex vs non-complex
  implementations.
\item
  there is little difference between the outlet of the expression as all
  modalities were rated above average.
\item
  There is little difference between the ratings for the affective
  impact on the robot and on the human who interacts with the robot.
\end{itemize}

\section{Applying the findings}

Based on the results, we can state that each affective expression
modality of the robot successfully conveyed an understanding of being
scolded by the sender in the interaction. The high ratings on all
modalities as seen in Figure \ref{graph_all} show that even low
complexity implementations such as status lamps can have a large
affective impact as a response to incoming stimuli. It should be noted
that since the participants were a culturally homogeneous group, and the
results are dependent on the cultural and social attributes of the
users, it may be difficult to achieve similar results with a group from
a different social or cultural background.

As some of the modalities were limited by physical aspects that were
hard to circumvent (E.g. a physical morphological response and larger
gestures are difficult to expand and enlarge beyond the physical
limitations of a small-sized robot), it could seem unfair to compare the
ratings of each modality. However, the results from the scolding
experiment indicate that normal intuition regarding the hierarchy of
expression modalities may not apply to this context. It is interesting
that the ratings in Figure \ref{graph_all} are quite high and close to
each other on all modalities despite differences in the complexity of
their implementations. As previously stated, the physical and contextual
limitations made some of the expression modalities stand out as being
very simplistic. E.g. the morphological expression modality consisted of
a blinking status led strip. Even that modality was rated as highly
impactful by the participants and shows that the robot managed to
generate some form of empathy despite it not being optimally shaped as
outlined in Riek et al.~\autocite{Riek2009HowAA}.

Some modalities also naturally attract more attention than others.
Anthropomorphic features such as eyes are high saliency objects that
immediately grab attention in an interaction such as outlined by
Birmingham et al.~2009 \autocite{Birmingham2009SaliencyDN}. With the
full focus of the test participants, one might assume that this modality
would receive a relatively higher rating than other lesser pronounced
features. However, this is not evident in the resulting data. These
counterintuitive results can be seen as an effect gained from the
high-intensity context in which the interaction takes place - that by
scolding the robot, people feel remorseful and in that state disregard
the complexity, size, and type of the response. This indicates that
there might be opportunities in aligning the engineering aspects of the
affective expression features of the robot with the emotional aspects of
the context. By considering the intensity of the interaction during the
design phase of affective robots and by dynamically aligning the
expression modalities with the measured intensity level of interaction,
it may be possible to expand the general expression abilities of future
affective robots.

We chose to use a simple non-humanoid robot in these tests. Using a
different type of robot could sway the results in a different direction.
Using a humanoid robot could perhaps make it easier to make test
participants view the robot from an anthropomorphic angle. This makes
the findings even more interesting as the robot we used did not rely on
cuteness or human features to convey affective information
\autocite{Caudwell2019WhatDH,Dubal2011HumanBS,Zhang2018AnalysisOI}. Our
initial assumption was that the difference in affective impact on the
robots and the affective impact of the humans would be large. However,
as these two affective states were rated relatively similar (in Fig.
\ref{graph_all}), it can be argued that for conflict scenarios it does
not matter where the emotion is interpreted as residing.

Some participants refused to interpret the robot in an anthropomorphic
manner and focused strictly on the technical aspects of the robot
instead. E.g. some participants stated that the robot could not make an
emotional impression on them as they viewed the robot solely as a
physical construction consisting of a black square box with lights
attached to it. However, high-mid to high ratings were given by these
participants despite them refusing to interpret the robot from an
anthropomorphic angle. They still considered the interaction as
affectively impactful but for different reasons. This is a topic we did
not investigate further in our tests. However, this may be an
opportunity for further studies in Human-Robot conflict Interaction
scenarios.

\section{Conclusion}

This paper investigated the potential ability of robots to defuse
conflict interactions using five different affective expression
modalities to convey acknowledgment of a received scolding. This was
accomplished by modifying the expressive behaviors and technical
expressive implementations of an affective robot designed for the
purpose. The results showed that all modalities were usable in the
context and that they worked to similar effect. Although the
implementation of several modalities was relatively simplistic, there
were no major measurable drawbacks in their rated affective impact. The
ratings were also similar for the affective impact interpreted from the
robot behavior and the impact on the humans in the interaction. The
robots managed to convey an emotional impact from the interaction and
the results indicate that defusing a conflict interaction may be
feasible simply by detecting the intensity and reacting with any
available expression modality. The results further highlight the effect
of placing humans and robots in high-intensity interaction scenarios and
indicate that the context and objective of the interaction may be a
viable catalyst for enforcing an anthropomorphic interpretation of robot
behaviors.

\balance
\bibliography{bibliography}
\bibliographystyle{IEEEtran}

\end{document}